# FRACTAL GEOMETRY OF LITERATURE: FIRST ATTEMPT TO SHAKESPEARE'S WORKS

Ali Eftekhari

*Electrochemical Research Center, P. O. Box 19395-5139, Tehran, Iran*

**ABSTRACT**

It was demonstrated that there is a geometrical order in the structure of literature. Fractal geometry as a modern mathematical approach and a new geometrical viewpoint on natural objects including both processes and structures was employed for analysis of literature. As the first study, the works of William Shakespeare were chosen as the most important items in western literature. By counting the number of letters applied in a manuscript, it is possible to study the whole manuscript statistically. A novel method based on basic assumption of fractal geometry was proposed for the calculation of fractal dimensions of the literature. The results were compared with Zipf's law. Zipf's law was successfully used for letters instead of words. Two new concepts namely Zipf's dimension and Zipf's order were also introduced. It was found that changes of both fractal dimension and Zipf's dimension are similar and dependent on the manuscript length. Interestingly, direct plotting the data obtained in semi-logarithmic and logarithmic forms also led to a power-law.

# 1. INTRODUCTION

Since the revolutionary discovery of fractal geometry by Mandelbrot[1], a numerous studies in different branches of science have been devoted to understand fractality of different objects and processes. There are several differences between fractal geometry and Euclidean geometry as the following ones. Firstly, the recognition of fractals is very modern (from such new classification point of view), as they have only been formally studied in the past two decades, whereas Euclidean geometry goes back over 2000 years. Secondly, whereas Euclidean shapes normally have a few characteristic sizes or length scales (e.g. the radius of a circle or the length of a side of a cube) fractals have so characteristic sizes. Fractal shapes are self-similar and independent of size or scaling. Thirdly, Euclidean geometry just provides a good description of manmade objects whereas fractals can be used as representation of naturally occurring geometries. It is likely that this limitation of our traditional language of shape is responsible for the sticking difference between mass produced objects and natural shapes. Finally, Euclidean geometries are defined by algebraic formulae. Fractals are normally the result of an iterative or recursive construction or algorithm.

It has been reported that most objects in nature and manmade items are fractal on both microscopic and macroscopic scales[2-5]. Indeed, importance and interesting feature of fractal geometry is related to its ability for analyzing natural structures and even different processes. Therefore, fractal analyses have been used in different branches of science. The most vivid works of fractals can be found in chemistry[6-12], as it is related to three different typical features. In this context, fractals have found at chemically reactive surfaces, molecular structures and even during a chemical reaction. The first and second features are similar to usual items of fractal, as they have been observed at fractal structure of surfaces. The latter feature is related to fractal processes, as suggests that chemical reactions may have fractal dimensions.

Fractals are of interest in art as well as science. This is due to both artistic and scientific investigations of art. Creating every fractal shapes can be included in art and *vice versa*. The appearance of fractality processes is also accompanied by a deep history in art. It comes from the First Art, music, which although is a manmade object but it can be extended to natural phenomena. Since the discovery of fractal geometry, the question of whether music has fractal geometry has been debated[13-17]. Hsu and Hsu have reported a method for the calculation of fractal dimension of music and have found a positive answer for this question[18]. They have recognized an inverse log-log relationship between the frequency and intensity of natural events[18-20]. Based on analyzed melody in terms of the interval between successive pitches, they have used the following formula:

$$F = c / i^D \qquad (1)$$

where $D$ is the fractal dimension of the composition, $i$ is the interval between two successive pitches, $F$ is the percentage frequency of $i$ and c is a constant proportionality factor. This method has been successfully used for analyzing and calculating the fractal dimension of different types of music. On the other hand, this is also an interesting subject area in research from both scientific and artistic points of view. Several types of music have been analyzed to find their fractal structures[18-22], and several music items have been created fractally[23-26] based on basic assumptions of fractal geometry.

In the context of the present paper, we would like to introduce fractal for another objects, which is of interest from both scientific and artistic points of view. It can be considered as a well-defined approach, as it is based on simple assumptions of fractal geometry and those have been elaborated for music. According to similarity of music and literature from both art and structure points of view, we used the above-mentioned method for the analysis of literature. It is simply possible, as letters in literature are completely equal to notes in music. Thus, we can use the mentioned equation for calculating the fractal dimension of literature. For keeping the original form of the equation, we use the same symbols with different meanings. Therefore, $D$ is the fractal dimension of the literature, $i$ is the interval between two letters in alphabetical series, $F$ is the percentage of $i$ and c is still a proportionality factor.

The problem, which should be taken into account, is the description of $i$ for literature. For music, $i$ is interval between notes, a typical note is chosen as the base note to calculate the value of $i$ for other notes in respect with the base note. Thus, the base note has $i = 0$. We modify this approach for alphabetical letters by introducing a assumptive letter before $A$ with incidence of zero. By choosing this assumptive letter as the base letter to calculate the values of $i$ for other letters (indeed, all letters). Now, for the base letter, which has no role in data analyzing, ($i = 0$) and for A ($i = 1$), …, and for Z ($i = 26$). The advantages of this modification is that the value of $i$ for the letters is equal to their ranks in alphabetical series. It should also be emphasized as it is aimed to study literature, it is necessary to obey from literature rules to use alphabetical order for the letters utilized in literature. This makes us able to compare the results obtained from fractal analysis of literature with those obtained from other statistical methods.

## 2. RESULTS

*2.1. Fractal Analysis*

Hamlet as a famous tragedy of Shakespeare was selected as a typical example. As expected, the appearance of different letters in literature has chaotic arrangement. This is observable in Fig. 1, where the number of the appearance of each letter is given for the Hamlet manuscript. The characteristic data for the mentioned literature are presented in Table 1. According to Eq.(1), fractal dimension of the literature can be determined from slope of the *F vs.* $1/i$ curve plotted in a log-log scale. The corresponding plot is illustrated in Fig. 2. Although, the data are dispersed, but slope of the curve based on mean square root approach can be determined. Consequently, the slope of the curve suggests a fractal dimension of 0.45 for the Hamlet manuscript.

As seen, the fitting of the curve is weak and has a low correlation coefficient ($R^2$) of 0.06712, far from a well-defined fitting data. This is in consistent with dispersing data in the curve. It goes back to the limitation of letters in literature structure. Of course, this is usual for this type of fractal structures. In fractal music, the fitting the data obtained from notes is also weak in comparison with usual curves in scientific works.

*2.2. Degree of Fractality*

Two different features have been elaborated from the concepts of fractal geometry[1]: *(i)* the objects are more complicated than that to be defined with Euclidean geometry (with simple integer dimensions) and they have non-integer dimensions depending to their complexity, *(ii)* the complex objects can be defined by terms of self-similarity or self-affinity. We know various fractal models and real objects, which have simple patterns of self-similarity or self-affinity. However, most of real fractal objects do not obey from a well-defined fractal pattern. Indeed, the revolutionary feature of fractal geometry is the first above-mentioned feature, which made it a universal theory. It is obvious that all objects are subject of the first feature of fractal geometry. In fact, it is very hard to find an object with an integer dimension. For example, it is difficult to find a completely surface without any roughness (even in microscopic scale), describing an integer dimension of 2.

There is a question that what is the difference between fractal and non-fractal surfaces? The question can be simply answered by introducing fractality factor $\zeta$, which is a dimensionless factor between 0 and 1 (can also be presented in %) to calculate how much the surface is defined by the fractal patterns. To understand the physical meaning of $\zeta$, we describe it for a classic fractal object with a known fractal pattern of Sierpinski gasket (Fig. 3a). Sierpinski gasket is one of the most famous fractal patterns with a known fractal dimension, $\log(3)/\log(2) = 1.58496$[1]. According to the Sierpinski gasket pattern, all four gaskets displayed in Fig. 5 have the same fractal dimension of 1.58496. But do they have the same fractality? If an artificial object fabricated based on the Sierpinski gasket has ideal pattern of the right gasket, the object can be exhausted and changed to the left gasket as the result of time progress or any other destroying factors such as corrosion. Although, the fractal dimension of all gaskets are the same, however, they lose their fractality from right to left, which can be expressed by decreasing $\zeta$.

Therefore, the value of $\zeta$ indicates the degree of fractality for an object. For known pattern of fractal the value of $\zeta$ approaches unity (100% fractality), but this value is lesser for real objects. When we design a fractal object like Sierpinski gasket theoretically on paper, the value of $\zeta$ is 1 suggesting 100% fractality of our mathematical pattern. However, if we fabricate this pattern on a gold surface (i.e., the conventional method for the fabrication of artificial fractal electrodes), we never have 100% fractality. Although, gold surface is very smooth, the roughness factor (the ratio of real surface area to geometrical area) for very smooth gold surface is about 1.2. Consequently, roughness of the own gold surface changes our desired fractality. As shown for the investigation of a series of real and artificial gold electrodes by electrochemical methods[27], the correlation coefficient of fitting data can be considered as a factor of the surface fractality or the value of $\zeta$.

This is also true for the system under investigation. Indeed, low correlation coefficient reported indicates weak fractality of the Hamlet manuscript. It is not strange, as we cannot expect strong fractality for the literature. If we take into account the case of music, which is very similar to the case under consideration, it can be expressed that the fractally generated music with a computerized approach have strong fractality with a $\xi$ approaching 1. Whereas, although the common kinds of music also have fractal structure but with weak fractality. Indeed, we can consider a fractally created music to have a well-defined fractal pattern like Sierpinski gasket as presented at the right side of Fig. 3, and a conventional music can be considered as that one illustrated at the left side. Similarly, we can consider a manuscript (e.g., Shakespeare's works under investigation) as a fractal object like the Sierpinski gasket presented at the left (in Fig. 3). On the contrary to fractally created music, we have no fractally created manuscript to be considered with a well-defined fractal pattern (such as Sierpinski gasket, as presented at the right side in Fig. 3). Bringing up the problem of fractality was to describe that weak data fitting for fractal analysis of the literature is understandable and the fractal dimension calculated for the Hamlet manuscript is of interest as well as those of other fractal objects.

It should also be emphasized that the value estimated for the fractal dimension of the Hamlet manuscript is not the purpose of this study. This is the first paper in this subject to show the possibility of the method proposed for mathematical study of literature. The method is reliable, as it is based on simple assumption of fractal geometry and as noted above the hypothesis was compared with music, which is a well-known example of fractals. The results obtained from analysis of different tragedies of Shakespeare are summarized in Table 2. Although, the values estimated for the fractal dimensions of different manuscripts are in the same range, however, there are significant differences between them. It is known that the value of fractal dimension can be used for comparative studies of similar objects. For example, if we prepare some gold surfaces by the deposition of gold at different condition, the fractal dimensions of such different Au-deposits are an excellent factor for comparison of the surface structures[28].

*2.3. Zipf's Law*

As stated above, the results are not satisfactory in comparison with other fractals we know. It is useful to compare the fractal analysis of the Shakespeare's works with statistical analysis based on other approaches. Unfortunately, there is no extensive method for the analysis of literature as we employed for statistical analysis of letters. The methods in this context are related to analysis of words rather than letters. The most famous method for this purpose is based on Zipf's law, which was proposed by George Kingsley Zipf.[29] Zipf's law suggests that frequency of occurrence of some event (*P*), as a function of the rank (*i*) when the rank is determined by the above frequency of occurrence, is a power-law function $P_i \sim 1/i^a$ with the exponent *a* close to unity. Similarity of this relation to that noted in Eq.(1) is obvious. Zipf's law has been widely used for the statistical analysis of texts[30-36]. The most famous example in this context is the statistical analysis of the most-repeating words such as "the", "of", "to", etc. Now, we examine the Zipf's law based on the strategy of this study to analyze letters of whole manuscript. The results are shown in Fig. 4 for the Hamlet manuscript. The data are similar to those presented in Fig. 1, however, as the letters are ordered in accordance with their frequency of occurrence, the changes are monotonic. The letters indicated on the curve show the arrangement of letters in accordance with their frequencies of occurrence to obey from the Zipf's law. We call this arrangement as the Zipf's order. If we transform this curve (Fig. 4) into a log-log scale (similar to that done in Fig 2), we can estimate the value of *a* noted in the Zipf's law, which is called Zipf's dimension $D_Z$. The results obtained for different tragedies of Shakespeare are summarized in Table 3. The required Zipf's orders for different tragedies are also illustrated in Fig. 5.

As seen, the Zipf's orders for different tragedies are very similar, and it suggests that the Zipf's law can be used as a general method for the statistical analysis of literature based on letters, as the frequency of each letter is approximately constant in English language. Indeed, this suggests that increase of the letter frequencies obey from a power-law. On the other hand, the data presented in Table 3 suggests an ideal behavior of the result accompanied by good correlation coefficients (when the letters are ordered with Zipf's order). Although, it is clear that the Zipf's dimension is different from the fractal dimension estimated previously (reported in Table 2), it can be concluded by comparison of the data presented in Table 2 and Table 3 that both dimensions proposed for the literature are commensurate. This means that the fractal dimension of a manuscript (as investigated for Shakespeare's works) is higher than other when its Zipf's dimension is higher than others, and *vice versa*. As illustrated in Fig. 6, the changes of both fractal dimension and Zipf's dimension are the same. In the other words, both fractal dimension and Zipf's dimension can be equivalently used for comparative investigation of different manuscripts.

Interestingly, it is obvious that the dimension (both Zipf and fractal ones) of the literature is dependent on the manuscript length. For the ten tragedies of Shakespeare under investigation, the dimension is higher for the shorter manuscripts. Although, there is an exception at about 100,000 letters-manuscript, this behavior is also applicable for shorter manuscripts. The equivalency of fractal dimension and Zipf's dimension indicates that though the literature have weak fractality, however, have certain structure of fractality, as Zipf's law is a known approach for the analysis of texts.

*2.4. Direct Data Plotting*

Both above-mentioned approaches for analysis of the literature were based on reciprocal power-laws. To use this type of power-law, $1/i$ or $1/n$ were used. Now, it is appropriate to make a direct statistical analysis of the literature. Once again, we use Hamlet manuscript as a typical example. We use different forms of semi-logarithmic and logarithmic data plotting. Figure 7 depicts the results obtained from $P$ *vs.* log ($i$), log ($P$) *vs.* $i$, and log ($P$) *vs.* log ($i$). To find the best method for data plotting, the correlation coefficients for the methods are compared. The values obtained for correlation coefficients of the three methods were 0.675249, 0.846831 and 0.954612, respectively. Similar results were also obtained for the other works of Shakespeare.

Interestingly, the data plotting based on full logarithmic (log-log) form provided the best result. This suggests that power-law function is an appropriate approach for analysis of literature and even has a better result in comparison with other statistical methods. This power-law function is similar to Zipf's law. As using n or 1/n leads to similar results were the data are plotted in a full logarithmic (log-log) form, the data plotting illustrated in Fig. 7(c) can be considered as a kind of Zipf's law. Indeed, the results obtained are indicative of the fact that Zipf's law is an appropriate statistical methods for analysis of literature based on alphabetical letters.

## 3. CONCLUSION

It was demonstrated that statistical analysis of letters based on fractal geometry or Zipf's law can be used to investigate literature. Two novel features were proposed in this study for the first time: *(i)* it was shown that literature could have fractality and *(ii)* the usefulness and powerfulness of Zipf's law for analysis of literature based on all letters employed. In this regard, the degree of fractality and new concepts such as Zipf's order and Zipf's dimension were also described. In conclusion, this proposes a new strategy for the analysis of written materials. On the contrary of the available method, which are based on the analysis of words and are just applicable for texts with meaning in a certain language, the method proposed can be used as a general approach even for random texts. In addition, the relationship of fractality and Zipf's law was also shown.

Table 1. Characteristics data for letters applied in the Hamlet manuscript.

| Letter | Letter interval ($i$) | Incidence | Incidence % ($F$) |
| --- | --- | --- | --- |
| *Base letter*[a] | 0 | 0 | 0 |
| *A* | 1 | 10,251 | 7.5931645963423 |
| *B* | 2 | 1,816 | 1.34515529284534 |
| *C* | 3 | 2,840 | 2.10365695577135 |
| *D* | 4 | 5,375 | 3.98139300608135 |
| *E* | 5 | 15,845 | 11.7367762197877 |
| *F* | 6 | 2,712 | 2.0088442479056 |
| *G* | 7 | 2,493 | 1.84662563054154 |
| *H* | 8 | 8,639 | 6.399117056658 |
| *I* | 9 | 8,905 | 6.59614971519152 |
| *J* | 10 | 110 | 0.0814796708221299 |
| *K* | 11 | 1,257 | 0.931090420212884 |
| *L* | 12 | 6,489 | 4.80655985422546 |
| *M* | 13 | 4,239 | 3.1399302237728 |
| *N* | 14 | 8,578 | 6.35393287556573 |
| *O* | 15 | 11,450 | 8.48129300830352 |
| *P* | 16 | 2,006 | 1.48589290608357 |
| *Q* | 17 | 218 | 0.161477893083857 |
| *R* | 18 | 8,100 | 5.99986666962956 |
| *S* | 19 | 8,668 | 6.42059806078384 |
| *T* | 20 | 12,450 | 9.2220172885047 |
| *U* | 21 | 4,738 | 3.50955163959319 |
| *V* | 22 | 1,219 | 0.902942897565239 |
| *W* | 23 | 3,110 | 2.30365251142567 |
| *X* | 24 | 177 | 0.131108197595609 |
| *Y* | 25 | 3,198 | 2.36883624808338 |
| *Z* | 26 | 120 | 0.0888869136241417 |

[a] This is an assumptive character introduced for calculating the value of I for 26 letters of English alphabets. All intervals were estimated in accordance with this assumptive letter, but it has no role in the statistical analysis of the literature.

Table 2. Calculated values for the fractal dimensions of different tragedies of Shakespeare.

| Tragedies | Number of total letters | Fractal dimension ($D_f$) | Correlation coefficient ($R^2$) |
|---|---|---|---|
| *Anthony and Cleopatra* | 116,209 | 0.5516 | 0.07923 |
| *Coriolanus* | 124,626 | 0.4707 | 0.06009 |
| *Hamlet* | 135,003 | 0.4500 | 0.06712 |
| *Julius Caesar* | 86,659 | 0.5269 | 0.07023 |
| *King Lear* | 115,986 | 0.5598 | 0.07934 |
| *Macbeth* | 77,524 | 0.5985 | 0.09261 |
| *Othello* | 115,245 | 0.5699 | 0.08031 |
| *Romeo and Juliet* | 105,834 | 0.5496 | 0.08686 |
| *Timon of Athenes* | 83,500 | 0.5358 | 0.07680 |
| *Titus Andranicus* | 92,467 | 0.5180 | 0.06743 |

Table 3. Analysis of different Shakespeare's tragedies based on Zipf's law.

| Tragedies | Zipf slope [a] | Correlation coefficient ($R^2$) | Zipf dimension [b] ($D_Z$) | Correlation coefficient ($R^2$) |
|---|---|---|---|---|
| *Anthony and Cleopatra* | 0.40587 | 0.921137 | 1.9320 | 0.972011 |
| *Coriolanus* | 0.40876 | 0.927299 | 1.8764 | 0.955026 |
| *Hamlet* | 0.40585 | 0.928312 | 1.6973 | 0.954612 |
| *Julius Caesar* | 0.40596 | 0.948025 | 1.9448 | 0.956628 |
| *King Lear* | 0.40276 | 0.913496 | 1.9424 | 0.955115 |
| *Macbeth* | 0.40247 | 0.928162 | 1.9414 | 0.974379 |
| *Othello* | 0.41361 | 0.925255 | 1.9720 | 0.961605 |
| *Romeo and Juliet* | 0.40588 | 0.920359 | 1.8441 | 0.977931 |
| *Timon of Athenes* | 0.41199 | 0.927070 | 1.8935 | 0.959367 |
| *Titus Andranicus* | 0.40869 | 0.936204 | 1.9558 | 0.961202 |

[a] Slope of the *F* (percent of incidence) – Letters (arranged in Zipf's order) plot.

[b] Estimated from the slope of the log (F) – log (1/*n*) plot, where *n* refers to the letter in Zipf's order. Indeed, $D_Z$ is the exponent of a in the Zipf's equation.

Figure Legends:

**Fig. 1.** Appearance of letters in a typical Shakespeare's work, a famous tragedy; Hamlet.

**Fig. 2.** Plotting $F$ against $1/i$ in a log-log scale for a typical Shakespeare's work, the Hamlet manuscript.

**Fig. 3.** Degree of fractality presented for a series of Sierpinski gaskets.

**Fig. 4.** The numbers of each letters repeated in the Hamlet tragedy as a function of letters ordered due to their intervals in Zipf's order.

**Fig. 5.** Zipf's order of letters for different Shakespeare's tragedies.

**Fig. 6.** Variations of both Zipf's dimension $D_Z$ and fractal dimension $D_F$ (fractal dimension was illustrated one unit higher as $D_F+1$ for comparison purpose) for different manuscripts. The manuscript# is referred to different Shakespeare's works investigated in this research arranged in accordance with their lengths (number of letters).

**Fig. 7.** Plotting the data obtained from Hamlet manuscript in different forms: (a) $P$ – log ($i$), (b) log ($P$) – $i$, (c) log ($P$) – log($i$). The percent of incidence ($P$) was obtained from the data reported in Table 1 and the rank $i$ is the Zipf's order reported in Fig. 5.

Figure 1

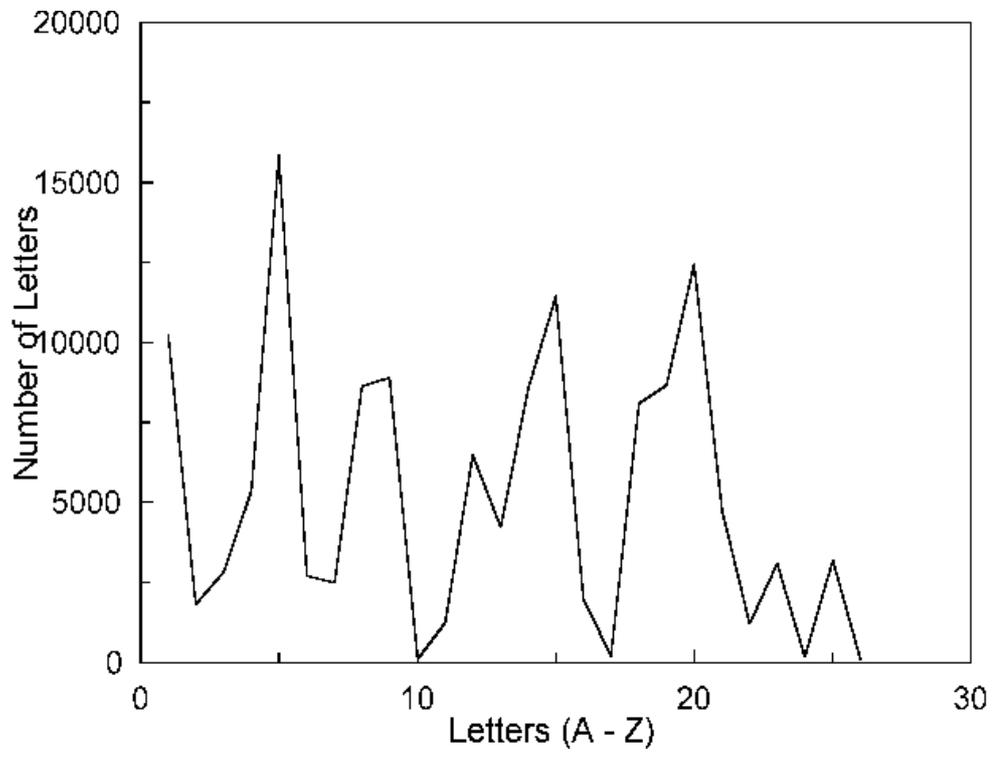

Figure 2

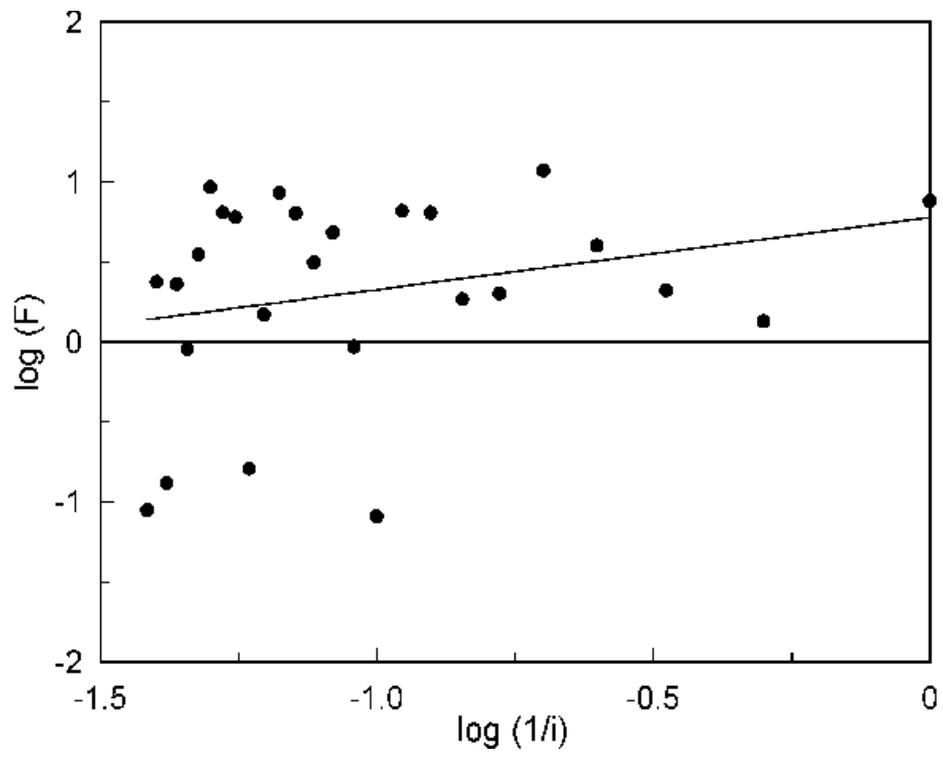

Figure 3.

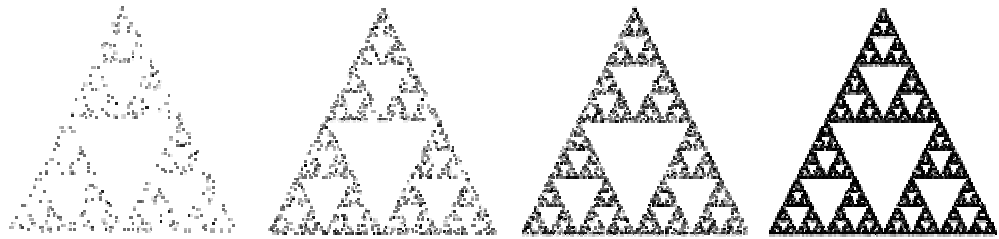

Figure 4.

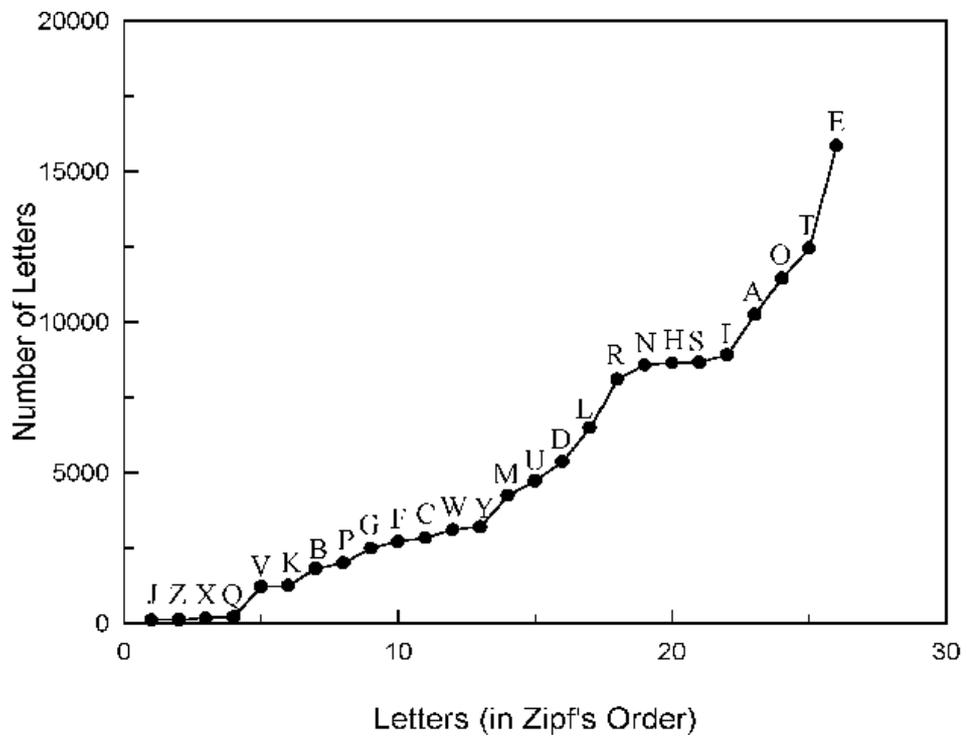

Figure 5.

| | |
|---|---|
| *Anthony and Cleopatra:* | Z J Q X K V B G F P W Y C M U D L H N I R S O T A E |
| *Coriolanus:* | Q J X Z K V P G B F Y W C M D L U H R N S A I O T E |
| *Hamlet:* | J Z X Q V K B P G F C W Y M U D L R N H S I A O T E |
| *Julius Caesar:* | Q J X Z K V P G B F W Y C M D L U H N R I S O A T E |
| *King Lear:* | Z Q J X V K B P F C G W Y M U D L I H S R N A O T E |
| *Macbeth:* | Z J X Q V K P G B Y F W C M U L D R I S N H A O T E |
| *Othello:* | Z Q J X K V P B G F C W Y M U D L R N H S I A T O E |
| *Romeo and Juliet:* | Z Q X J K V P B G F C W Y M U D L N R S H I A O T E |
| *Timon of Athenes:* | Z Q J X K V B G P C F Y W M U D L R H N I S A O T E |
| *Titus Andranicus:* | Z X J Q K V P G B F C Y W M L D U H I R N S A O T E |

Figure 6.

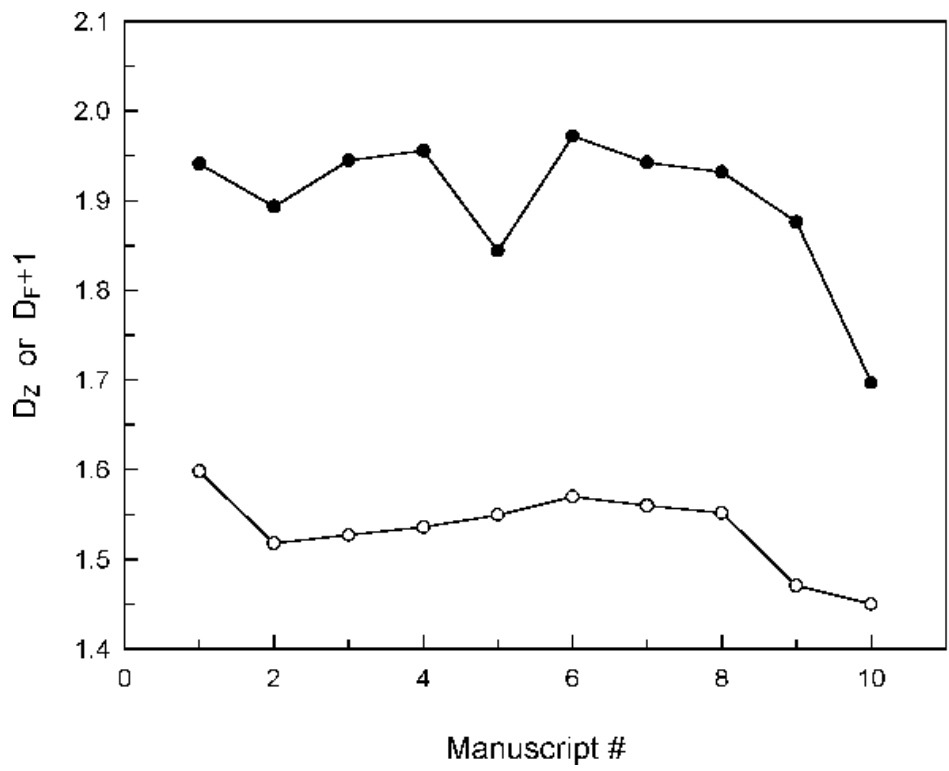

Figure 7(a).

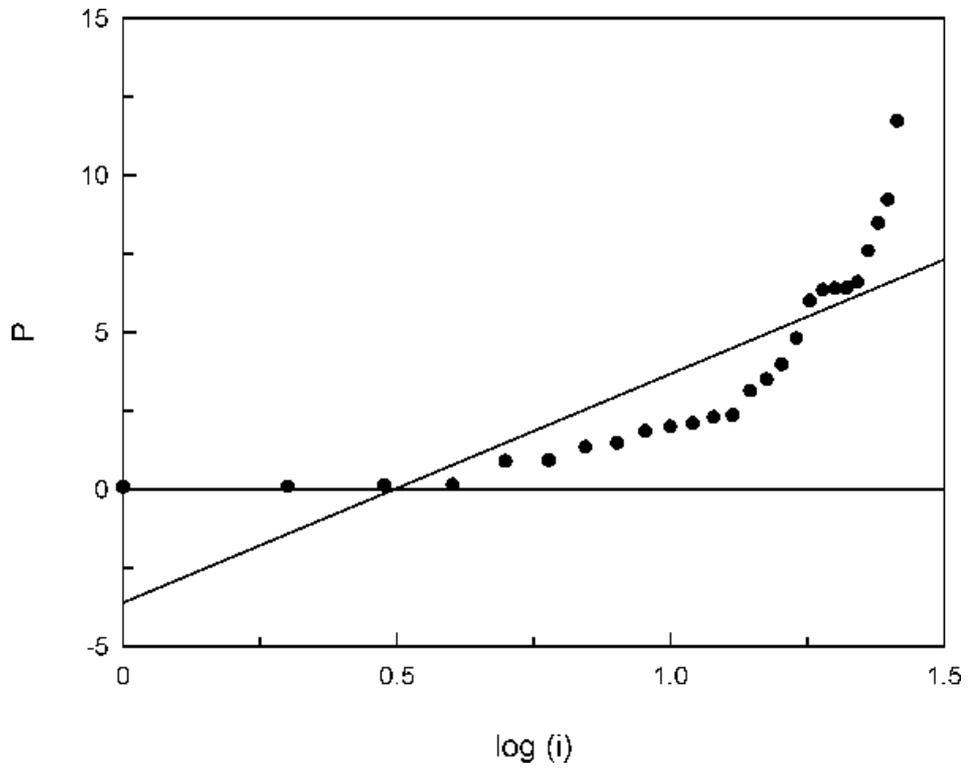

Figure 7(b).

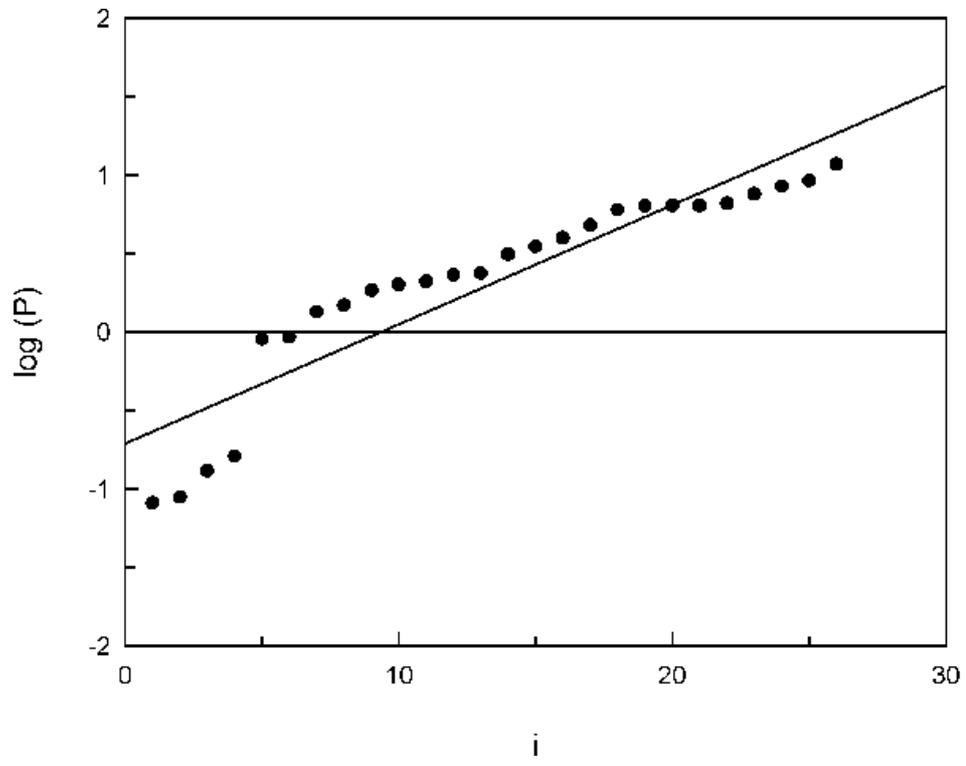

Figure 7(c).

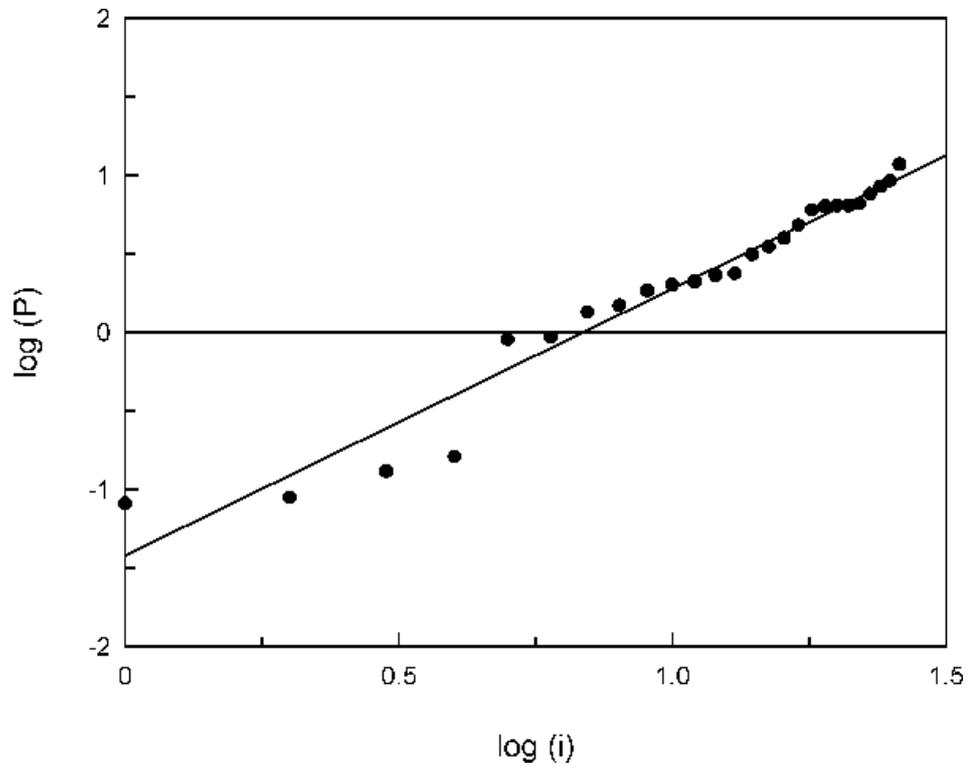